\documentclass{article}

\PassOptionsToPackage{numbers,compress}{natbib}

 \usepackage[preprint]{neurips_2026}

\usepackage[utf8]{inputenc} % allow utf-8 input
\usepackage[T1]{fontenc}    % use 8-bit T1 fonts
\usepackage[pagebackref,breaklinks,colorlinks,citecolor=teal]{hyperref}
\usepackage{url}            % simple URL typesetting
\usepackage{booktabs}       % professional-quality tables
\usepackage{amsfonts}       % blackboard math symbols
\usepackage{nicefrac}       % compact symbols for 1/2, etc.
\usepackage{microtype}      % microtypography
\usepackage{xcolor}         % colors
\usepackage{makecell} % in preamble
\usepackage{multirow}
\usepackage{graphicx}
\usepackage{wrapfig}
\usepackage{float}
\usepackage{colortbl}
\usepackage[table]{xcolor}
\usepackage{amsmath} 
\usepackage{arydshln}
\usepackage{xcolor}
\definecolor{darkgreen}{RGB}{0, 150, 0}
\usepackage{subcaption}
\usepackage{pifont}

\newcommand{\xmark}{\ding{55}}

\title{BRepCLIP: Contrastive Multimodal Pretraining on BRep Primitives for CAD Understanding}

\begin{document}

\def\thefootnote{*}\footnotetext{These authors jointly supervised this work.}
\def\thefootnote{$\dagger$}\footnotetext{Corresponding author}
\author{
  \begin{tabular}[t]{c}
    Muhammad Usama$^{1,2}$ \\
    \texttt{muhammad.usama@dfki.de}
  \end{tabular} \hspace{1em}
  \begin{tabular}[t]{c}
    \textbf{Didier Stricker}$^{1,2}$ \\
    \texttt{didier.stricker@dfki.de}
  \end{tabular} \hspace{1em} \\
  \\
  \begin{tabular}[t]{c}
    \textbf{Mohammad Sadil Khan}$^{1,2*\dagger}$ \\
    \texttt{mohammad.khan@dfki.de}
  \end{tabular} \hspace{1em}
  \begin{tabular}[t]{c}
    \textbf{Muhammad Zeshan Afzal}$^{2*}$ \\
    \texttt{muhammad.zeshan.afzal@dfki.de}
  \end{tabular} \\
  \\
 \begin{tabular}[t]{c}
    $^{1}$\text{DFKI, Germany} \\
    % \texttt{didier.stricker@dfki.de}
  \end{tabular} \hspace{0.2em}
  \begin{tabular}[t]{c}
    $^2$\text{RPTU Kaiserslautern-Landau, Germany} \\
    % \texttt{sk.aziz.ali@dfki.de}
  \end{tabular} \hspace{0.2em}
}

\maketitle

\begin{abstract}

Learning representations of CAD models is a largely open problem. While 3D representation learning has flourished around point clouds and meshes, the native format of CAD - boundary representations (BReps), which encodes exact parametric surfaces, curves, and their topology, has received little attention as a representation learning substrate. We introduce \textbf{BRepCLIP}, the first framework to align BRep geometry with language and image embeddings through contrastive pretraining. We model each CAD object as a sequence of face and edge tokens with separate discrete vocabularies for surface and curve geometry, augmented with spatial and semantic descriptors that capture surface types (e.g., cylindrical, torus, NURBS) and curve primitives (e.g., line, arc, B-spline). A transformer encoder aggregates these tokens into a global BRep embedding, aligned with CLIP's text and image encoders via a joint contrastive objective. BRepCLIP generates more discriminative and semantically grounded embeddings than existing point-based alternatives, improving Top-1 retrieval over OpenShape by \textbf{40.4\%}, \textbf{22.0\%}, and \textbf{23.9\%} on ABC, CADParser, and Automate, respectively, and improving zero-shot classification on FabWave by \textbf{15\%} in Top-1 score. We further demonstrate its utility as a CAD-aware similarity metric for evaluating text- and image-conditioned CAD generation, establishing the importance of structure-aware pretraining for multimodal CAD understanding. Project page is available at \href{}{https://muhammadusama100.github.io/BrepClip2026/}

\end{abstract}

\section{Introduction}

Computer-aided design (CAD) is the backbone of modern engineering, underpinning the design of everything from consumer electronics to aerospace components~\cite{Brown2009CADDC,cad-challenges}. CAD models are represented as a BRep structure, which provides exact, parametric descriptions of geometry organized into faces, edges, and their topological adjacencies~\cite{Brepnet}. Unlike generic 3D assets, BRep geometry is precise by construction. Every surface has an analytic type, every edge has a defined curve, and the topology encodes how parts connect and bind one another. In practice, engineers rarely design from scratch. They search large internal repositories to find and reuse existing parts, adapting them to new specifications. This process, known as CAD retrieval, is central to reducing design time, avoiding redundant modeling, and ensuring manufacturing consistency across product lines~\cite{survey}. Despite its industrial importance, learning general-purpose representations that support open-vocabulary CAD retrieval remains a largely open problem. 

% While 3D representation learning has advanced rapidly~\cite{gao2023mixcon3d}, driven by progress on point clouds, meshes, and implicit surfaces, virtually none of this progress has been directed at the CAD domain.

\begin{wrapfigure}{r}{0.48\textwidth}
    \centering
    \vspace{-\intextsep} % remove top gap
    \includegraphics[width=\linewidth]{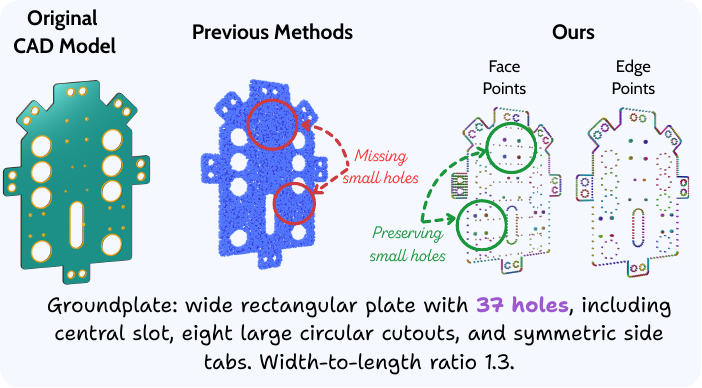}
    \vspace{-3pt} % tighten space below image
    \caption{Compared to point clouds, our BRep-aware representations (edge, face points) preserve both geometry and fine-grained structures (e.g., holes, rounded corners) for accurate CAD representation learning.}
    \label{fig:prev-method}
    \vspace{-\intextsep} % remove bottom gap
\end{wrapfigure}

Existing multimodal 3D alignment methods~\cite{xue2024ulip,liu2023openshape} learn powerful joint representations of point clouds, images, and text, demonstrating strong performance on generic 3D object understanding. However, these methods are fundamentally designed around point cloud representations and cannot be directly applied to CAD models without first discarding their native BRep structure. Converting a BRep to a point cloud reduces a precisely structured boundary representation to an unordered set of coordinates, erasing the analytic surface types, curve primitives, and topological adjacency that are intrinsic to CAD geometry. For generic 3D assets, this approximation may be acceptable, but for CAD, it is a fundamental information loss. The geometric features most critical for engineering interpretation, such as small holes, chamfered and filleted edges, sharp boundaries, face-to-face adjacency, and exact surface curvature, are precisely what point clouds fail to encode (Figure ~\ref{fig:prev-method}). A representation that cannot distinguish a cylindrical bore from a planar pocket, or a filleted edge from a sharp one, cannot support fine-grained CAD retrieval or reliable generation evaluation.

We introduce BRepCLIP, the first contrastive representation learning framework to operate directly on BRep primitives.
Each CAD model is represented as a set of BRep face and edge primitives, where each primitive is encoded through sampled local geometry together with its semantic type and topological grouping. We learn discrete surface and curve tokens from these faces and edge points using a dVAE model~\cite{yu2022point}. In addition to spatial descriptors, these tokens also encode semantic descriptors.
% 
% Rather than converting CAD geometry into a lossy intermediate, BRepCLIP treats faces and edges as first-class tokens. Each CAD model is represented as a sequence of face and edge tokens with separate discrete vocabularies for surface and curve geometry, augmented with semantic descriptors encoding primitive type and spatial descriptors encoding geometric position. 
A transformer aggregates these tokens into BRep-aware tokens, which are aligned with frozen CLIP text and image encoders via a symmetric contrastive objective. 

% The central insight is that BRep structure is itself a supervision signal: by preserving the native decomposition of CAD geometry into typed, semantically meaningful primitives, the model learns embeddings that are grounded in the actual engineering content of the shape.

Operating directly on BReps presents a core challenge: unlike point clouds, BReps have no canonical ordering, vary in the number of faces and edges across models, and contain heterogeneous geometry types such as planes, cylinders, tori, NURBS surfaces, or lines, arcs, and B-Splines, within a single model. We address this with a hybrid dual-DVAE tokenization scheme, training separate discrete autoencoders for faces and edges to produce dedicated codebooks for surface and curve geometry. This prevents geometrically dissimilar primitives from sharing a vocabulary and allows each branch to specialize to its own geometric domain. Each token is further augmented with semantic descriptors derived from primitive type, so the transformer reasons over typed CAD entities rather than anonymous geometric patches.

We evaluate BRepCLIP on three tasks. On text-to-CAD retrieval, BRepCLIP outperforms all point-based baselines on ABC, CADParser, and Automate. On zero-shot CAD classification, we transfer directly to FabWave~\cite{fabwave} without fine-tuning, again exceeding point-cloud counterparts. Finally, we introduce BRepCLIP-Score, a geometry-aware metric for evaluating text- and image-conditioned CAD generation, and show it correlates more reliably with human expert judgments than CLIP score~\cite{clip} or Chamfer Distance. Our contributions are as follows:

\begin{itemize}
    \item \textbf{First contrastive} representation learning framework operating natively on BRep primitives, bridging CAD geometry with language and image modalities.
    
\item \textbf{Hybrid dual-dVAE} tokenization with separate discrete tokens for face and edge geometry, enabling semantically typed tokenization of heterogeneous BRep primitives.

\item \textbf{State-of-the-art results} on text-to-CAD retrieval and zero-shot CAD classification.

\item \textbf{BRepCLIP score}, a new CAD-aware similarity metric for evaluating text- and image-conditioned CAD generation, validated against human expert judgments.

\end{itemize}

% we showcase that BRepCLIP is a reliable metric for benchmarking text or image-conditioned models compared on CLIP-Score or .

% we introduce a CAD generation evaluation benchmark comprising outputs from six recent text-to-CAD models, and validate the BRepCLIP score as a geometry-aware automatic metric against the CLIP score and the Point-BERT score, using a human expert study with five CAD design professionals as ground truth.

\section{Related Work}

\noindent \textbf{3D representation learning and CAD.}
3D representation learning has progressed from PointNet's hierarchical point aggregation~\cite{qi2017pointnet,qi2017pointnet++} to transformer-based self-supervised pretraining with Point-BERT~\cite{yu2022point}, and more recently to multimodal alignment. ULIP~\cite{xue2023ulip} aligned point clouds, images, and text against a frozen CLIP space, ULIP-2~\cite{xue2024ulip} scaled this through automatic caption generation from rendered views, and OpenShape~\cite{liu2023openshape} pushed further with multi-dataset ensembling and stronger backbones for open-world recognition. Despite their strength in generic 3D assets, these methods are ill-suited for CAD retrieval. CAD retrieval is not a coarse semantic matching problem. It requires discriminating between shapes that look globally similar but differ in the details that matter most for engineering: a threaded hole versus a smooth bore, a chamfered edge versus a fillet, an extruded pocket versus a boss. Point clouds reduce geometry to an unordered set of surface samples, discarding the analytic surface types, curve primitives, and topological adjacency that are intrinsic to CAD~\cite{jayaraman2021uv,Brepnet}. This structural information is erased at the point of conversion. No downstream architecture can recover what was discarded at the input.

Recognizing this, a line of work has moved toward learning directly on BRep structure. BReps are the native format of CAD models, organizing geometry into typed faces such as planes, cylinders, tori, and NURBS surfaces, and typed edges such as lines, arcs, and B-splines, connected through an explicit topological graph. This structure is not incidental. It is the primary carrier of engineering semantics. UV-Net introduced UV-domain surface sampling with graph-based topology learning~\cite{jayaraman2021uv}, while BRepNet exploited native BRep connectivity through message passing over faces, edges, and coedges~\cite{Brepnet}. BRep-BERT applied masked modeling over BRep subgraphs using a GNN tokenizer~\cite{lou2023brep}, and BRT brought attention-based encoding to boundary representations~\cite{Zou_2025}. MultiCAD proposed contrastive representation learning between point clouds and CAD sequences~\cite{ma2023multicad}, and BrepCoder aligned BRep geometry with structured CAD code for multi-task reasoning~\cite{kim2026brepcoder}. However, all of these methods target recognition, segmentation~\cite{sharp2023,brepdetnet}, reconstruction~\cite{cadsignet}, or within-CAD structural pretraining. None learn language or image-aligned representations over native BRep primitives, a prerequisite for open-vocabulary retrieval that BRepCLIP is the first to address.

\noindent \textbf{CAD retrieval, generation, and evaluation.}
CAD retrieval is a practically critical but surprisingly underexplored problem. In engineering workflows, retrieval enables part reuse, design search, and manufacturing planning. These tasks demand fine-grained geometric discrimination rather than coarse object-level similarity. As surveyed in~\cite{cadretrieval,survey}, learning-based CAD retrieval has largely relied on shape signatures, voxel descriptors, or rendered silhouettes, none of which capture the topological and parametric richness of BReps. Early work on scan-to-CAD retrieval focused on aligning clean CAD models to noisy RGB-D scans~\cite{scan2cad}, and FastCAD extended this to real-time retrieval and alignment using contrastive shape embeddings~\cite{fastcad}. However, both operate on generic 3D representations and target scene-level alignment rather than language-driven engineering retrieval. Jones et al. proposed self-supervised pretraining directly on BRep geometry using a hybrid implicit/explicit surface representation, demonstrating strong few-shot transfer on BRep benchmarks~\cite{selfsupervised}. Yet this work focuses on within-CAD recognition tasks and does not align BRep geometry with language or image modalities. OSCAR studied open-set CAD retrieval from language and image prompts~\cite{pulli2026oscar}, and CAD-RAG introduced a retrieval-augmented generation framework combining multiple modalities~\cite{ananthakrishnan2025multi}. However, both operate on non-native representations and are not designed for large-scale contrastive pretraining over BRep structure. The recent release of CADCAP-1M from DreamCAD~\cite{dreamcad} is the largest CAD captioning dataset to date and finally makes large-scale multimodal BRep representation learning tractable. BRepCLIP is the first method to exploit it through native BRep pretraining.

The dominant direction in multimodal CAD research has meanwhile been generation. DeepCAD established the sequence-modeling view of parametric CAD~\cite{deepcad}, and subsequent work extended this to reconstruction and generation from point clouds, BReps, text, and images~\cite{liu2024point2cad,zhang2024brep2seq,text2cad,xu2024cad,li2025cad,nurbgen,wang2025cad,marvel-40m+,chen2025cadreview}. Yet as generative CAD has grown, evaluation has not kept pace. Generated models are typically assessed with Chamfer Distance or CLIP score. These metrics are borrowed from point cloud and vision-language literature and are blind to BRep structure. A model that produces the correct overall silhouette but wrong surface topology, missing holes, or incorrect edge types will score well on these metrics while failing every engineering criterion that matters. BRepCLIP-Score addresses this directly. It is a CAD-aware similarity metric grounded in BRep embeddings, validated against human expert judgments on outputs from six recent text-to-CAD models.
\section{BRepCLIP Architecture}

We present BRepCLIP, a multimodal CAD representation learning framework that aligns native BRep geometry with text and images through contrastive pretraining. Unlike generic multimodal 3D encoders built on point clouds, BRepCLIP operates directly on CAD primitives and treats faces and edges as first-class entities throughout the pipeline. Each CAD model is represented by face ($G_f$) and edge ($G_e$) point sets together with primitive-type semantics. We tokenize these primitives with separate face and edge tokenizers, producing dedicated discrete tokens for surface and curve geometry. The resulting face-edge token sequence is then enriched with spatial and semantic cues and processed by a transformer encoder, whose learnable \texttt{[CLS]} token yields a global BRep embedding for multimodal alignment.

\subsection{Hybrid Face-Edge Tokenization}

\begin{wrapfigure}{r}{0.58\textwidth}
    \centering
    \vspace{-\intextsep} % remove top gap
    \includegraphics[width=\linewidth]{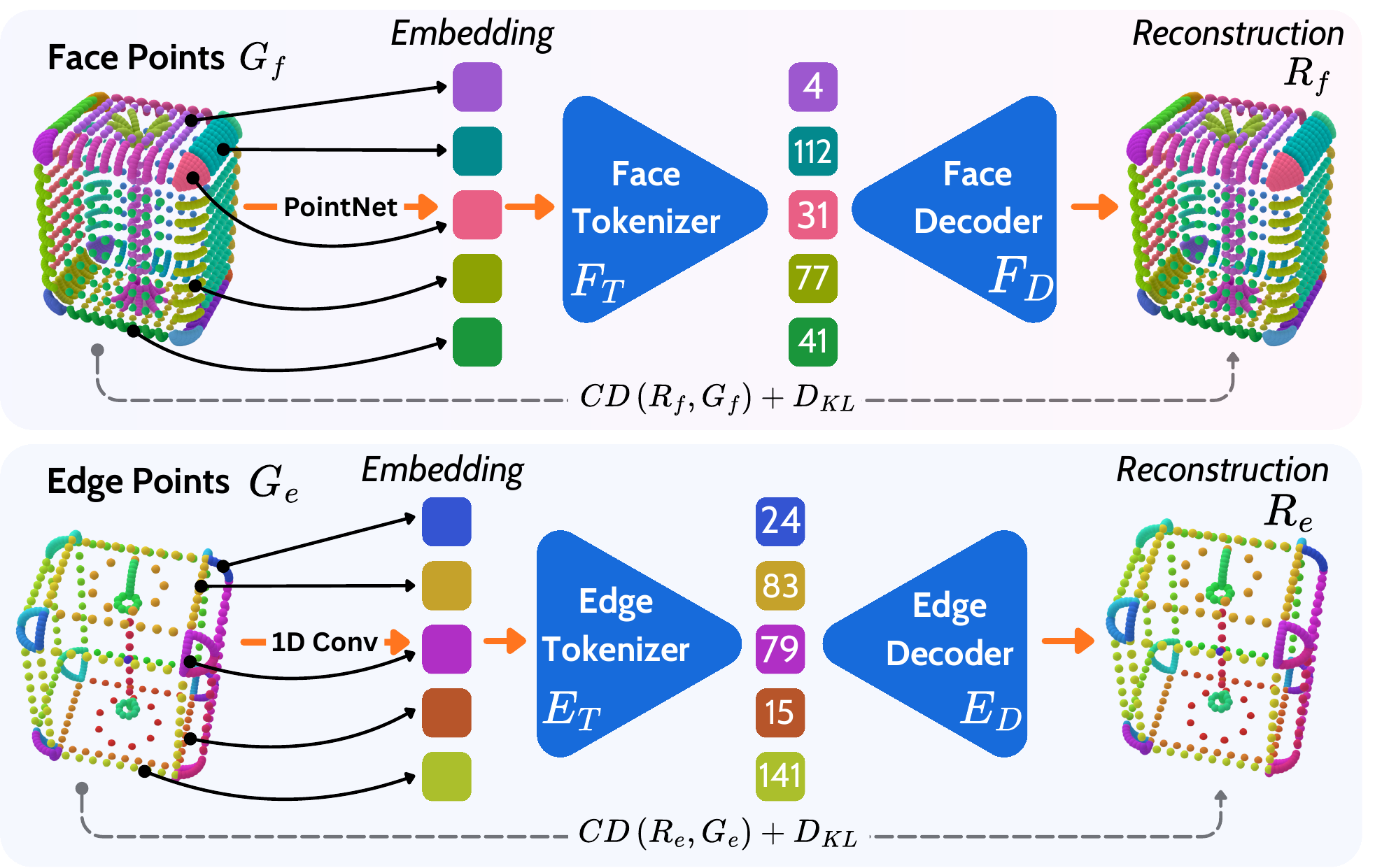}
    \vspace{-10pt} % tighten space below image
    \caption{\textbf{Hybrid dual-dVAE tokenization.} Face and edge points are tokenized independently using separate discrete VAEs with dedicated codebooks.} 
    \label{fig:brep-encoder}
    \vspace{-\intextsep} % remove bottom gap
\end{wrapfigure}

We encode BRep geometry through a tokenization scheme over faces and edges as shown in Figure~\ref{fig:brep-encoder}. Unlike Point-BERT \cite{yu2022point}, which uses local point neighborhoods to group points and generate tokens, we instead use corresponding face and edge segmentation to group points semantically.
% 
% forms tokens by grouping local point neighborhoods, we directly use native BRep primitives as tokenization units.
% 
We train two separate dVAEs for faces and edges. The face dVAE uses a PointNet-style encoder with face tokenizer $F_T$ and folding-based decoder $F_D$ to reconstruct surface geometry. The edge dVAE uses a lightweight 1D convolutional encoder with edge tokenizer $E_T$ and decoder $E_D$ to reconstruct ordered curve geometry. Separate codebooks are essential as faces and edges exhibit fundamentally different geometric structures. Each dVAE is trained by minimizing a reconstruction loss with a KL regularization term as
\vspace{-.5\baselineskip}
\begin{equation}
    \mathcal{L}_{x} = CD(R_x,G_x) + D_{KL},
\end{equation}
where $x$ can be either face or edge, and $CD(R_x, G_x)$ denotes the Chamfer Distance between sampled points from the reconstructed and ground-truth geometry, and $D_{KL}$ is the KL divergence regularizing the discrete latent space~\cite{yu2022point}.

\subsection{Structure-Aware Global BRep Encoding}
After obtaining discrete face tokens $F_T(G_f)$ and edge tokens $E_T(G_e)$, we construct a unified BRep sequence by concatenating them with a learnable \texttt{[CLS]} token as
% \vspace{.\baselineskip}
% \begin{equation}
% \mathbf{Z} = [\texttt{CLS} \; ; \; F_T(G_f) + f_g + f_m + f_s + f_d \; ; \; E_T(G_e) + e_m + e_s + e_d]
% \end{equation}
\begin{equation}
\mathbf{Z^B} = [\texttt{CLS} \; ; \; F_T(G_f) + f_m + f_s + f_d \; ; \; E_T(G_e) + e_m + e_s + e_d]
\end{equation}
where $f_m$ and $e_m$ are modality indicators distinguishing faces from edges, $f_s$ and $e_s$ are spatial descriptors derived from primitive centroids, and $f_d$ and $e_d$ are semantic descriptors encoding primitive type. The geometry and modality terms form the content embedding of each token, while the spatial and semantic terms form its positional embedding. This sequence is processed by a transformer encoder, and the final \texttt{[CLS]} representation serves as the global BRep embedding, capturing both 3D structure and fine-grained surface and curve semantics.

% encode them as a unified BRep sequence with a transformer backbone. Each sequence element combines discrete geometry with CAD-specific metadata that describes the role of the underlying primitive in the boundary representation.

% Concretely, each sequence element is described by four signals: a discrete geometry token, a face/edge modality indicator, a spatial descriptor derived from the primitive centroid, and a semantic descriptor encoding primitive attributes. In the model, the geometry and modality signals are summed to form the token content embedding, while the spatial and semantic signals are summed to form the positional/meta embedding. This allows the encoder to reason jointly over primitive geometry, primitive type, spatial location, and CAD-specific semantics.

% The resulting sequence is processed by a transformer encoder with a learnable \texttt{[CLS]} token for global aggregation. Compared with standard Point-BERT \cite{yu2022point} style encoding, which relies primarily on local geometric patches and generic positional cues, our design explicitly incorporates BRep structure and primitive semantics, making it better suited for fine-grained CAD representation learning.

\subsection{Multimodal Contrastive Alignment}

\begin{figure}[t]
    \centering
    \includegraphics[width=1\linewidth]{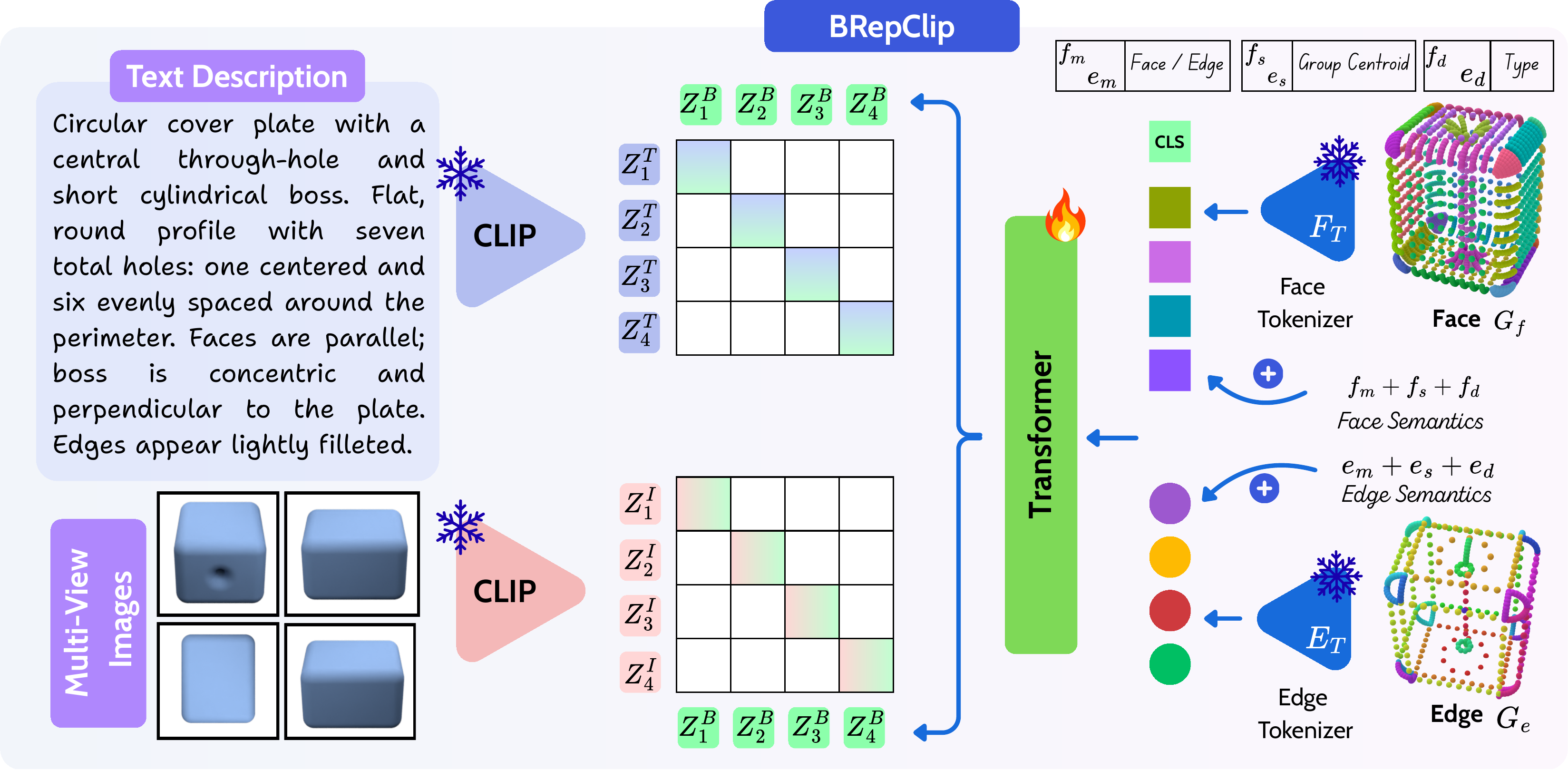}
    \caption{\textbf{BRepCLIP.} Face and edge point sets, $G_f$ and $G_e$, are tokenized by frozen face ($F_T$) and edge ($E_T$) tokenizers and encoded by a transformer with modality, spatial, and semantic cues to produce a global BRep embedding. Frozen CLIP text and image encoders provide caption and multi-view image embeddings for BRep--text and BRep--image contrastive training.}
    \label{fig:placeholder}
\end{figure}

BRepCLIP aligns BRep geometry with text and image modalities through a three-branch contrastive framework consisting of a structure-aware BRep encoder, a frozen CLIP text encoder, and a frozen CLIP image encoder~\cite{clip}. The BRep branch encodes the face-edge token sequence with a transformer encoder, producing a global shape embedding $\mathbf{Z}^B \in \mathbb{R}^d$ from the \texttt{[CLS]} token via a lightweight MLP projection head. In parallel, the frozen CLIP text and image encoders produce embeddings $\mathbf{Z}^T$ and $\mathbf{Z}^I$ respectively, each projected into the same shared latent space. Only the BRep branch is trained; the text and image encoders remain frozen throughout. Training is driven by two symmetric InfoNCE contrastive objectives: a BRep-text loss $\mathcal{L}_{bt}$ and a BRep-image loss $\mathcal{L}_{bi}$. For a batch of $N$ matched CAD-text pairs, $\mathcal{L}_{bt}$ is defined as:

\vspace{-10pt}
\begin{equation}
\mathcal{L}_{bt} = -\frac{1}{2N} \sum_{i=1}^{N} \left[ \log \frac{\exp(\mathbf{Z}^B_i \cdot \mathbf{Z}^T_i / \tau)}{\sum_{j=1}^{N} \exp(\mathbf{Z}^B_i \cdot \mathbf{Z}^T_j / \tau)} + \log \frac{\exp(\mathbf{Z}^T_i \cdot \mathbf{Z}^B_i / \tau)}{\sum_{j=1}^{N} \exp(\mathbf{Z}^T_i \cdot \mathbf{Z}^B_j / \tau)} \right]
\end{equation}

where $\tau$ is a learnable temperature parameter and all embeddings are $\ell_2$-normalized prior to similarity computation. The BRep-image loss $\mathcal{L}_{bi}$ is defined analogously by substituting $\mathbf{Z}^T$ with $\mathbf{Z}^I$. The total training objective is:
\vspace{-.5\baselineskip}
\begin{equation}
\mathcal{L} = \mathcal{L}_{bt} + \mathcal{L}_{bi}
\end{equation}

This design keeps the multimodal alignment framework simple and compatible with existing retrieval pipelines, while the BRep encoder learns representations grounded in both language semantics and visual appearance.

\section{Experiments}

\textbf{Datasets}. We primarily use CADCap-1M from DreamCAD \cite{dreamcad}, specifically its high-quality ABC subset, which provides CAD models paired with captions and multiview renderings from CADCap-1M~\cite{dreamcad}. From this subset, we use 400K samples for training and 10K for validation. These data are used to train both the primitive tokenizers and the full BRepCLIP model. For each sample, we extract a structured BRep from the STEP file using a PythonOCC pipeline extended from BRepNet \cite{Brepnet}. We also sample dense point clouds for point-based baselines and use the DreamCAD  multiview renderings for image supervision.

\textbf{Implementation}.
Training proceeds in two stages. In the first stage, we train separate dVAEs for faces and edges on 4 NVIDIA A100 GPUs using AdamW with cosine decay and warmup, and annealing schedules for both Gumbel-Softmax temperature and KL-divergence weight. The face dVAE is trained for 100 epochs with a codebook size of 8192, and the edge dVAE for 200 epochs with a codebook size of 2048, both with latent dimension 256. In the second stage, BRepCLIP is trained for 38 epochs on a single NVIDIA A100 GPU using AdamW with learning rate $10^{-3}$ and weight decay $0.05$. The BRep transformer encoder is a 12-layer transformer with hidden dimension 384 and 6 attention heads, projected to a 512-dimensional shared embedding space. We use frozen OpenCLIP ViT-bigG-14 encoders for text and image, and optimize a weighted sum of BRep--text and BRep--image contrastive losses with equal weights. Training uses mixed precision, gradient checkpointing, and gradient clipping with an effective batch size of 200.

% \textbf{Experimental Setup.} We use the ABC split from DreamCAD as the main training source, as it provides CAD models paired with captions and multiview renderings. We use 400K samples for training and 10K for validation. For each model, we extract a structured BRep representation and a point-cloud representation, both paired with the corresponding caption; multiview images are additionally used for multimodal baselines. To support large-scale BRep processing on ABC, we extend the BRepNet extraction pipeline to perform dataset-wide feature extraction. For all baselines, we follow the training configurations reported in their original papers whenever applicable, including encoder architecture, optimizer settings, learning-rate schedule, and contrastive batch size. When adaptation to CAD data is required, we keep the original recipe fixed and only replace the input representation and dataset. We evaluate the resulting representations on two downstream tasks: zero-shot retrieval and zero-shot classification.

\noindent\textbf{Experimental Setup.} We train on the ABC split from DreamCAD~\cite{dreamcad}, which provides 400K CAD models paired with captions and multiview renderings, with 10K samples held out for validation. For each model, we extract both a structured BRep representation and a point cloud, paired with the corresponding caption. Multiview images are additionally used for multimodal baselines. To support large-scale BRep processing, we extend the BRepNet~\cite{Brepnet} extraction pipeline to dataset-wide feature extraction. All baselines follow their original training configurations; when adaptation to CAD data is required, we keep the original recipe fixed and only replace the input representation and dataset. We evaluate on three downstream tasks: zero-shot text-to-CAD retrieval , zero-shot CAD classification and generative CAD evaluation.

\begin{figure}[t]
    \centering
    \includegraphics[width=1\linewidth]{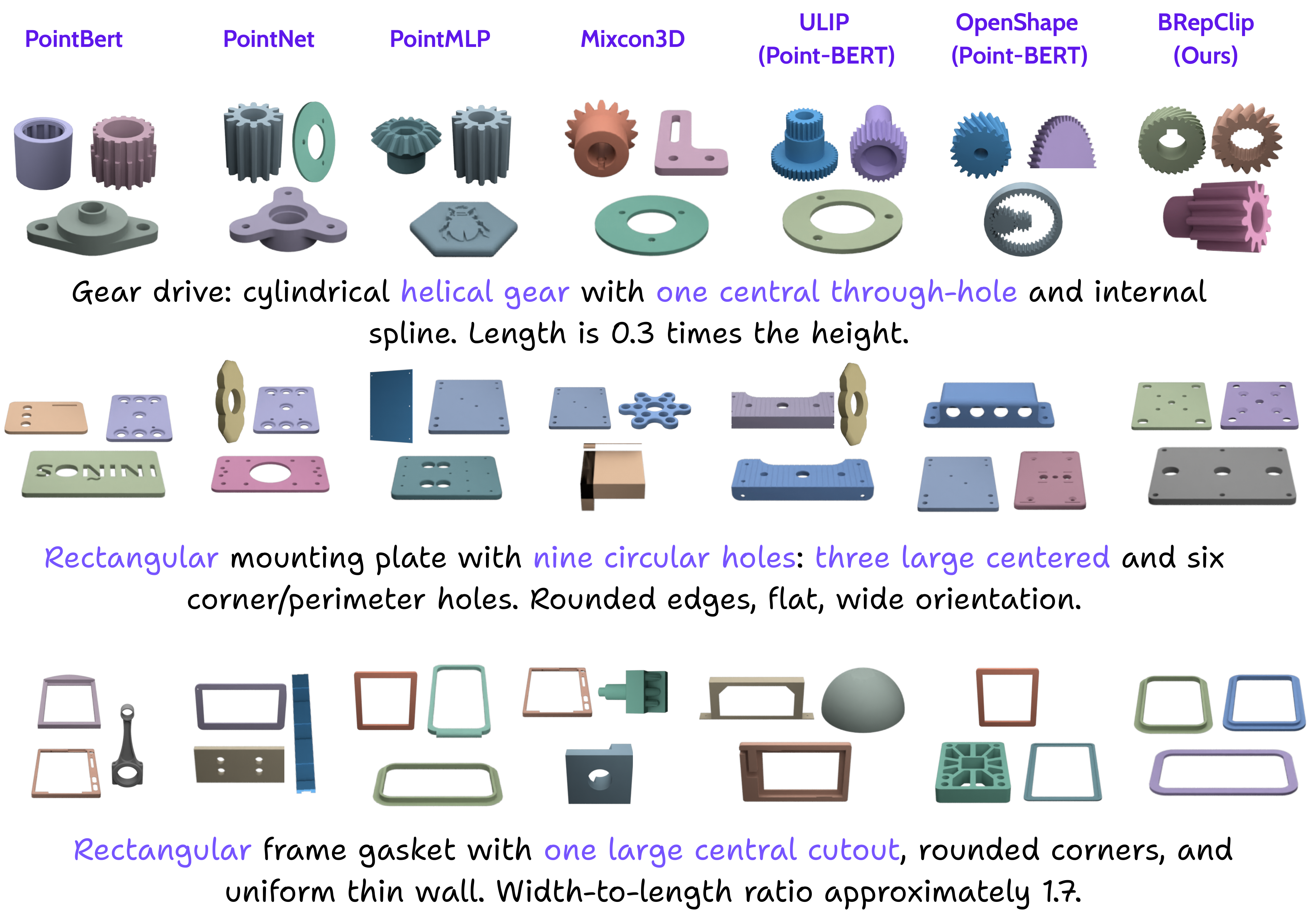}
    \caption{\textbf{Qualitative retrieval results.} Given a text query, BRepCLIP retrieves CAD models that faithfully match fine-grained geometric details such as hole count, edge topology, and surface type compared to Point-based baselines.}
    \label{fig:qualitative}
\end{figure}

\begin{table*}[t]
\def\arraystretch{1.25}%
\centering
\caption{Zero-shot text-to-CAD retrieval results across different CAD databases. Retrieval performance is reported using Top-$k$ accuracy and Chamfer Distance (CD). Chamfer Distance (CD) is scaled by $10^3$ }
\label{tab:retrieval_results}
\resizebox{\textwidth}{!}{
\begin{tabular}{l|ccccc|ccccc|ccccc}
\hline
\textbf{Method}
& \multicolumn{5}{c|}{\textbf{ABC}} 
& \multicolumn{5}{c|}{\textbf{CADParser}} 
& \multicolumn{5}{c}{\textbf{Automate}} \\
\cmidrule(lr){2-6}
\cmidrule(lr){7-11}
\cmidrule(lr){12-16}
& \textbf{Top-1} & \textbf{Top-5} & \textbf{Top-10} & \textbf{Top-20} & \textbf{CD} $\downarrow$
& \textbf{Top-1} & \textbf{Top-5} & \textbf{Top-10} & \textbf{Top-20} & \textbf{CD} $\downarrow$
& \textbf{Top-1} & \textbf{Top-5} & \textbf{Top-10} & \textbf{Top-20} & \textbf{CD} $\downarrow$ \\
\hline

Point-BERT \cite{yu2022point} & 2.60 & 9.36  & 15.80 & 22.72 & 61.56 & 1.10 & 4.40  & 7.00  & 10.90 & 45.12 & 0.91 & 3.55 & 6.04 & 9.64 & 71.58 \\
PointNet \cite{qi2017pointnet} & 3.31 & 12.07 & 19.38 & 29.60 & 62.27 & 0.40 & 1.90  & 3.50  & 6.40  & 64.80 & 3.33 & 10.72 & 16.41 & 23.96 & 68.13 \\
PointMLP \cite{pointmlp} & 0.90 & 3.50  & 6.00  & 9.50 & 68.43 & 1.10 & 3.70  & 6.70  & 9.30  & 54.55 & 1.02 & 3.79 & 6.39 & 10.20 & 75.51 \\
BRepEncoder  & 4.30 & 16.30 & 24.70 & 33.90 & 61.11 & 2.10 & 8.20  & 13.20 & 19.00 & 40.32 & 4.82 & 14.30 & 21.06 & 29.55 & 68.48 \\ \hdashline
MixCon3D \cite{gao2023mixcon3d} & 1.20 & 2.10  & 4.20  & 8.12 & 74.18 & 0.19 & 1.74  & 2.33  & 5.12  & 69.83 & 0.18 & 2.33 & 3.88 & 6.54 & 94.15 \\
ULIP \cite{xue2023ulip} & 2.30 & 4.00  & 7.40  & 12.20 & 63.48 & 0.70 & 2.90  & 4.60  & 7.10  & 67.33 & 0.92 & 3.11 & 5.06 & 7.95 & 91.41 \\
OpenShape \cite{liu2023openshape} & 6.12 & 18.17 & 24.88 & 34.36 & 71.63 & 4.10 & 13.40 & 19.70 & 29.30 & 43.33 & 7.60 & 19.86 & 27.58 & 36.45 & 79.82  \\
BRepCLIP   & \textbf{8.59}   & \textbf{24.52}    & \textbf{35.08}    & \textbf{47.89}  & \textbf{58.16} & \textbf{5.00} & \textbf{15.08} & \textbf{22.12}   & \textbf{30.60}    & \textbf{35.28}    & \textbf{9.42}    & \textbf{24.18} & \textbf{32.86} & \textbf{42.83} & \textbf{60.32} \\

\hline
\end{tabular}
}
\end{table*}
\subsection{Text-to-CAD Retrieval Task}

In the text-to-CAD retrieval task, the goal is to retrieve the most relevant CAD model from a gallery given a text query, using cosine similarity in the shared embedding space. We consider a zero-shot setting in which all gallery instances are unseen during training.

% We evaluate the learned representation on text-to-CAD retrieval. Given a text query, the model retrieves the most relevant CAD model from a gallery by ranking candidates according to cosine similarity in the shared embedding space. We consider a zero-shot setting, where all gallery instances are unseen during training.

\noindent\textbf{Task Dataset and Protocol.} All methods are trained on the same ABC split from CADCap-1M~\cite{dreamcad}, using 400K samples for training and 10K for validation. Retrieval is evaluated on three held-out datasets: a 91K held-out ABC split, CADParser~\cite{cadparser}, and Automate~\cite{automate}. The held-out ABC split serves as the in-domain retrieval benchmark, while CADParser and Automate are used for zero-shot retrieval transfer, since neither dataset is seen during training. For all datasets, BRep embeddings are precomputed offline for the gallery models. Concretely, the retrieval benchmarks contain 91K query model pairs for ABC, 40K for CADParser, and 65K for Automate, where each text query is evaluated against the full corresponding CAD gallery.

% \textbf{Gallery Dataset}. The retrieval gallery is constructed from three sources: a held-out ABC split from DreamCAD, CAD-Parser, and Automate. For each CAD model, we precompute its 3D embedding using the corresponding encoder, and retrieve nearest neighbors for each query in the joint embedding space.

\textbf{Baselines}. We compare against point-based 3D encoders (PointNet~\cite{qi2017pointnet}, PointMLP~\cite{pointmlp}, Point-BERT~\cite{yu2022point}), multimodal alignment frameworks (ULIP~\cite{xue2023ulip}, MixCon3D~\cite{gao2023mixcon3d}, OpenShape~\cite{liu2023openshape}), and our proposed BRep-based models. Our method is evaluated in two forms: \textit{BRepEncoder}, which uses our BRep-native encoder with text supervision only, and \textit{BRepCLIP}, which further adds image supervision. To ensure a fair comparison, all baselines are retrained on the same 400K ABC split used for BRepCLIP. We preserve the original training recipes of the respective methods whenever applicable, including encoder architecture, optimizer settings, learning-rate schedule, and contrastive batch size, and only replace the input representation and dataset where necessary.
% An important property of our approach is that the proposed BRep encoder is not restricted to the standalone BRepCLIP model. Since it produces a CAD-aware 3D representation, it can also replace generic point-based 3D backbones in existing multimodal frameworks. In our experiments, we therefore evaluate both the standalone BRepCLIP model and a hybrid variant in which the OpenShape 3D encoder is replaced by our BRep encoder. This allows us to test whether explicit BRep-aware 3D encoding improves retrieval within a strong multimodal architecture.

\noindent\textbf{Metrics.} We report Top-$k$ retrieval accuracy for $k \in \{1,5,10,20\}$ together with Chamfer Distance (CD). To measure geometric similarity beyond exact instance matching, we additionally compute CD on a random subset of 10K queries from each dataset. For each query, we retrieve the top-5 candidates, compute the Chamfer Distance between the ground-truth CAD model and each retrieved candidate, average over the top-5 retrieved results, and then average again over all query samples to obtain a single dataset-level CD score.

% \noindent\textbf{Results.} Table~\ref{tab:retrieval_results} shows that BRepCLIP achieves the strongest retrieval performance across all three datasets. It obtains the best Top-$k$ accuracy at every reported value of $k$ and also the lowest Chamfer Distance, indicating that the retrieved CAD models are both more semantically relevant and more geometrically similar to the ground-truth targets. On ABC, BRepCLIP improves Top-1 accuracy from 6.12 for OpenShape to 8.59 and reduces CD from 0.071 to 0.058. On CADParser, it reaches 5.00 Top-1 and 30.60 Top-20 with the best CD of 0.035. On Automate, it achieves 9.42 Top-1 and 42.83 Top-20, again with the best CD of 0.060. The text-only BRepEncoder already outperforms generic point-based encoders, showing the value of native BRep structure for CAD retrieval, while the full BRepCLIP further improves performance through multimodal alignment with both text and image supervision. Qualitative examples in Figure~\ref{fig:qualitative} show that BRepCLIP better captures fine-grained engineering properties such as hole count, edge topology, and surface type, whereas point-based baselines often retrieve only globally similar shapes.

\noindent\textbf{Results.} 
Table~\ref{tab:retrieval_results} shows that BRepCLIP achieves the strongest retrieval performance across all three datasets. It attains the best Top-$k$ accuracy at every reported value of $k$ and also the lowest Chamfer Distance, indicating that the retrieved CAD models are both more semantically relevant and more geometrically faithful to the ground-truth targets. On ABC, BRepCLIP improves Top-1 accuracy from 6.12 to 8.59, a relative gain of 40.4\%, while reducing CD from 0.071 to 0.058. On CADParser, it reaches 5.00 Top-1 and 30.60 Top-20, outperforming OpenShape by 22.0\% and 4.4\%, respectively, and achieves the best CD of 0.035. On Automate, BRepCLIP achieves 9.42 Top-1 and 42.83 Top-20, corresponding to relative improvements of 23.9\% and 17.5\% over OpenShape, while also lowering CD from 0.079 to 0.060. Notably, the text-only BRepEncoder already outperforms generic point-based encoders, confirming the importance of native BRep structure for CAD retrieval, and the full BRepCLIP further improves over it through multimodal alignment with both text and image supervision. Qualitative examples in Figure~\ref{fig:qualitative} further show that BRepCLIP better captures fine-grained engineering properties such as hole count, edge topology, and surface type, whereas point-based baselines often retrieve only globally similar shapes.

% These results demonstrate that explicitly modeling BRep structure yields more discriminative embeddings for fine-grained CAD retrieval.

% \textbf{Metrics}. We report Top-1, Top-5, Top-10, and Top-20 retrieval accuracy. To assess geometric fidelity beyond exact instance matching, we further report F1-score, Chamfer Distance (CD), and human evaluation (HE). We compare against generic 3D encoders and multimodal baselines to evaluate whether structured BRep-aware representations provide a stronger basis for CAD retrieval than conventional point-based embeddings.

% \textbf{Results}. As shown in Table~\ref{tab:retrieval_results}, BRepCLIP substantially improves retrieval performance over generic point-based encoders on the ABC test split. Moreover, replacing the original Point-BERT backbone in OpenShape with our BRep encoder further improves retrieval quality, suggesting that CAD-aware geometric encoding is complementary to broader multimodal supervision. Overall, these results indicate that explicitly modeling BRep structure yields more discriminative embeddings for fine-grained CAD retrieval.

\subsection{Zero-Shot Classification}

\begin{figure}[t]
    \centering
    \includegraphics[width=1\linewidth]{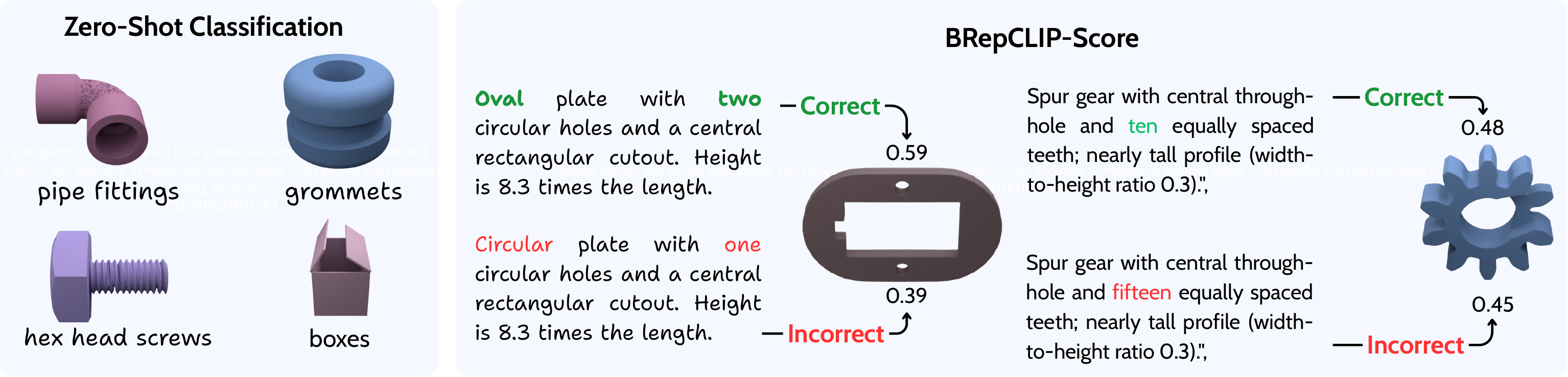}
\caption{Qualitative results for zero-shot classification and BRepCLIP-Score. \textbf{Left}: BRepCLIP supports zero-shot CAD classification via class-level text matching. \textbf{Right}: BRepCLIP-Score assigns higher similarity to prompt-faithful CAD outputs and lower similarity to mismatched ones.}
\label{fig:class-brepclip-score}
\end{figure}

% We evaluate the learned representation on zero-shot CAD classification using the FabWave dataset~\cite{fabwave}, which provides 39 predefined CAD categories after filtering incomplete samples. Unlike generic 3D recognition benchmarks, FabWave defines engineering-oriented classes and therefore offers a more relevant testbed for CAD understanding.

\noindent\textbf{Task.} Given a BRep embedding, we perform zero-shot CAD classification by matching each model against class-level text descriptors without any fine-tuning.

\noindent\textbf{Setup.} We evaluate on FabWave~\cite{fabwave}, which is not used during training and is treated as a zero-shot transfer benchmark for CAD classification. The original manifest contains 4,421 samples across 45 categories. After filtering out 43 broken or incomplete assets, the final benchmark contains 4,378 valid samples spanning 39 engineering-oriented categories. 
\begin{wraptable}{r}{0.44\textwidth}
\vspace{-1pt}
\def\arraystretch{1.15}
\centering
\caption{Zero-shot classification (FabWave)}
\label{tab:classification_results}
\vspace{2pt}
\resizebox{\linewidth}{!}{
\begin{tabular}{l|ccc}
\hline
\textbf{Method} & \textbf{Top-1} & \textbf{Top-5} & \textbf{Top-10} \\
\hline
Point-BERT \cite{yu2022point} & 17.34 & 40.21 & 56.04 \\
PointNet \cite{qi2017pointnet} & 15.74 & 38.78 & 54.37 \\
PointMLP \cite{pointmlp} & 18.80 & 41.00 & 59.02 \\
BRepEncoder & 21.81 & 43.40 & 60.74 \\ \hdashline
ULIP \cite{xue2023ulip} & 21.65 & 47.28 & 60.62 \\
MixCon3D \cite{gao2023mixcon3d} & 34.10 & 63.93 & 78.18 \\
OpenShape \cite{liu2023openshape} & 33.58 & 68.73 & 81.73 \\
BRepCLIP & \textbf{38.62} & \textbf{70.28} & \textbf{86.71} \\
\hline
\end{tabular}
}
\vspace{-20pt}
\end{wraptable}
All models are trained on ABC and transferred directly to FabWave without further fine-tuning. For evaluation, we define class-level text descriptors for the 39 valid categories and perform zero-shot classification by matching each CAD embedding to these class embeddings.

\noindent\textbf{Baselines and Metrics.} We compare against the same baselines as in retrieval and report Top-1, Top-5, and Top-10 accuracy.

\noindent\textbf{Results.} Table~\ref{tab:classification_results} shows that BRepCLIP achieves the best performance overall, reaching 38.62 Top-1, 70.28 Top-5, and 86.71 Top-10 accuracy. Among 3D-only encoders, BRepEncoder performs best, with 21.81 Top-1 accuracy, outperforming all point-based alternatives. These results indicate that CAD-aware BRep encoding yields more transferable semantic representations for zero-shot CAD classification.

\subsection{BRepCLIP Score for Generative CAD Evaluation}

\noindent\textbf{Motivation}. Evaluating text-to-CAD generation requires more than visual similarity. 
\begin{wrapfigure}{r}{0.5\textwidth}
    \centering
    \vspace{-\intextsep}
    \includegraphics[width=\linewidth]{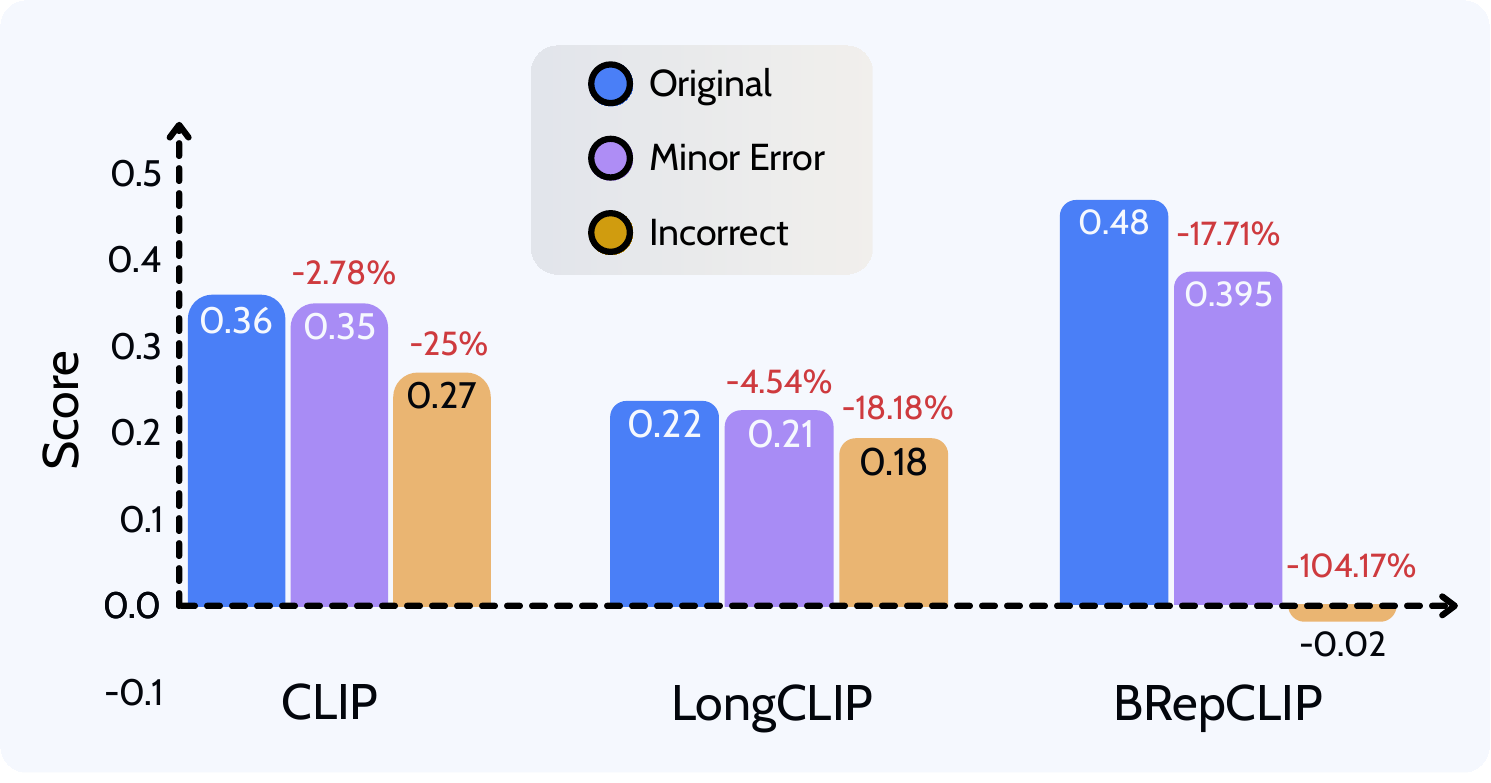}
    \vspace{-10pt}
    \caption{Score Sensitivity to prompt corruption.}
    \label{fig:score_clip_long_brep}
    \vspace{-\intextsep}
\end{wrapfigure}
A model that looks correct when rendered may still be missing holes, chamfers, or correct edge topology. CLIP score operates on 2D projections and cannot capture these details. Chamfer Distance measures global shape proximity but is insensitive to local topology. BRepCLIP-Score addresses both limitations by grounding evaluation directly in BRep embeddings, where surface types, edge primitives, and topological structure are explicitly represented.

\textbf{BRepCLIP-Score.} Given a text prompt $t$ and a generated CAD model $x$, we define
\begin{equation}
\mathrm{BRepCLIP\mbox{-}Score}(t,x)=\cos\!\big(f_{\text{text}}(t),\,f_{\text{3D}}(x)\big)
\end{equation}
where $f_{\text{text}}(t)$ is the text embedding of the prompt and $f_{\text{3D}}(x)$ is the BRep embedding produced by our encoder. To test sensitivity to semantic mismatch, we sample 10,000 CAD models from CADParser, Automate, and a held-out ABC split, and compare scores under three conditions: the original caption, a mildly corrupted GPT-generated caption, and a fully mismatched caption.

\textcolor{red}{}\textbf{Sensitivity to prompt corruption.} Figure~\ref{fig:score_clip_long_brep} shows that BRepCLIP-Score is substantially more sensitive to prompt corruption than image-based similarity metrics. 
\begin{wraptable}{r}{0.50\textwidth}
\vspace{-10pt}
\def\arraystretch{1.2}
\centering
\caption{Benchmarks of text-to-CAD methods.}
\vspace{3pt}
\label{tab:application2_scoring}
\resizebox{\linewidth}{!}{
\begin{tabular}{l|ccccc}
\hline
\textbf{Method}
& \cellcolor{orange!20}\textbf{CD} $\downarrow$
& \cellcolor{blue!10}\makecell{\textbf{CLIP} \\ \textbf{Score} $\uparrow$}
& \cellcolor{blue!10}\makecell{\textbf{Human} \\ \textbf{Score} $\uparrow$}
& \cellcolor{blue!10}\makecell{\textbf{GPT} \\ \textbf{Score} $\uparrow$}
& \cellcolor{blue!10}\makecell{\textbf{BRepCLIP} \\ \textbf{Score} $\uparrow$} \\
\hline
Ground Truth & - & 0.37 & 9.7 & 9.8 & 0.61 \\
\hdashline
DeepCAD \cite{deepcad} & 86.54 & 0.24 & 2.2 & 2.4 & 0.15 \\
Text2CAD \cite{text2cad}  & 86.54 & 0.26 & 3.6 & 3.5 & 0.16 \\
Cadrille \cite{cadrille} & 155.80 & 0.26 & 3.5 & 3.7 & 0.16 \\
Text2CQ (Q3B) \cite{text2QR} & 68.15 & \textbf{0.33} & 5.0 & 4.9 & 0.31 \\
Text2CQ (GL) \cite{text2QR}  & 71.27 & 0.32 & 4.6 & 4.5 & 0.25 \\
Text2CQ (CG) \cite{text2QR} & 77.91 & 0.31 & 4.1 & 3.9 & 0.22 \\
CADFusion \cite{CADFusion} & \textbf{56.36} & 0.29 & \textbf{5.5} & \textbf{5.8} & \textbf{0.35} \\
\hline
\end{tabular}
}
\vspace{-10pt}
\end{wraptable}
Under mild corruption, it drops by 17.71\%, compared with 2.78\% for CLIP score and 4.54\% for LongCLIP. Under full mismatch, the drop increases to 104.17\%, compared with 25.00\% and 18.18\%, respectively. This indicates that BRepCLIP-Score better reflects semantic inconsistencies that arise from incorrect geometry, rather than rewarding only visual resemblance.

% \noindent\textbf{BRepCLIP-Score.} Given a text prompt $t$ and a generated CAD model $x$, we define BRepCLIP-Score as
% \begin{equation}
% \mathrm{BRepCLIP\mbox{-}Score}(t,x)=\cos\!\big(f_{\text{text}}(t),\,f_{\text{3D}}(x)\big)
% \end{equation}
% where $f_{\text{3D}}$ is the BRep embedding produced by our encoder. To test sensitivity to semantic mismatch, we sample 10,000 CAD models from CADParser, Automate, and a held-out ABC split, and compare scores under three conditions: the original caption, a mildly corrupted GPT-generated caption, and a fully mismatched caption.

For benchmarking generative models, we evaluate on 15,000 examples from the ABC dataset using outputs from recent text-to-CAD methods, including DeepCAD~\cite{deepcad}, Text2CAD~\cite{text2cad}, CADRille~\cite{cadrille} Text2CQ~\cite{text2QR}, and CADFusion~\cite{CADFusion}. In addition to automatic metrics, we conduct both human and GPT-based evaluation following the protocol used in DreamCAD. Specifically, for each prompt, evaluators are shown multiview renderings of CAD generations from all competing methods together with the input text, and assign a score from 0 to 10 based on semantic similarity between the generated CAD model and the caption. Human evaluation is performed by five CAD designers, and the final human score is obtained by averaging their ratings. GPT evaluation uses the same multiview renderings and caption, and assigns scores under the same 0--10 semantic-faithfulness criterion. The resulting human and GPT scores are therefore preference-style measures of prompt faithfulness grounded in caption-to-geometry consistency rather than reconstruction accuracy alone. As shown in Table~\ref{tab:application2_scoring}, BRepCLIP-Score aligns more closely with both human and GPT judgments than CLIP score, indicating that it provides a more faithful evaluator of text-conditioned CAD generation quality.

\subsection{Ablation Study}
\begin{wraptable}{r}{0.44\textwidth}
\vspace{-12pt}
\def\arraystretch{1.15}
\centering
\caption{Ablation of BRepCLIP components on ABC retrieval.}
\vspace{4pt}
\label{tab:ablation_component}
\resizebox{\linewidth}{!}{
\begin{tabular}{l|cccc}
\hline
\textbf{Method} & \textbf{Top-1} & \textbf{Top-5} & \textbf{Top-10} & \textbf{Top-20} \\
\hline
Edge-only & 1.26 & 6.44 & 10.24 & 18.39 \\
Face-only & 3.40 & 13.12 & 19.24 & 26.39 \\
BRepCLIP & \textbf{8.59} & \textbf{24.52} & \textbf{35.08} & \textbf{47.89} \\
\hline
\end{tabular}
}
\end{wraptable}
\textbf{BRepCLIP Modality components.} We ablate the contribution of each BRep primitive branch by comparing edge-only, face-only, and full BRepCLIP variants. As shown in Table~\ref{tab:ablation_component}, both reduced variants perform substantially worse than the full model. Using only face primitives lowers Top-1 retrieval from 8.59 to 3.40, a drop of 60.4\%, while using only edge primitives further reduces it to 1.26, corresponding to an 85.3\% drop. Similar trends hold for Top-20, where the face-only and edge-only variants fall by 44.9\% and 61.6\%, respectively. These results confirm that surface and boundary geometry provide complementary cues, and that jointly encoding both is essential for discriminative CAD retrieval.

% These results show that neither surface geometry nor boundary geometry alone is sufficient for robust CAD retrieval. Instead, BRepCLIP benefits from jointly modeling faces and edges, which provide complementary cues about surface structure, local boundaries, and primitive transitions.
% \textbf{BRepCLIP components}. We analyze the contribution of different BRep primitive branches by comparing three variants: a face-only encoder, an edge-only encoder, and the full BRepCLIP model that jointly encodes both faces and edges.
% % \vspace{4pt}
% As shown in Table~\ref{tab:ablation_component}, both the face-only and edge-only variants perform substantially worse than the full BRepCLIP model. In particular, using only faces reduces Top-1 retrieval from 9.20 to 2.40, while using only edges further drops it to 1.26. This shows that neither surface geometry nor boundary geometry alone is sufficient for robust CAD retrieval. Instead, BRepCLIP benefits from jointly modeling faces and edges, which provide complementary information about surface structure, local boundaries, and primitive transitions.

\begin{wraptable}{r}{0.44\textwidth}
\vspace{-4pt}
\def\arraystretch{1.15}
\centering
\caption{Effect of batch size on BRepCLIP for ABC retrieval.}
\vspace{4pt}
\label{tab:ablation_batch_size}
\resizebox{\linewidth}{!}{
\begin{tabular}{c|cccc}
\hline
\textbf{Batch} & \textbf{Top-1} & \textbf{Top-5} & \textbf{Top-10} & \textbf{Top-20} \\
\hline
128 & 3.15 & 10.79 & 18.22 & 28.42 \\
200 & 8.59 & 24.52 & 35.08 & 47.89 \\
400 & \textbf{8.61} & \textbf{24.53} & \textbf{35.11} & \textbf{47.90} \\
\hline
\end{tabular}
}
\end{wraptable}

\noindent\textbf{Batch Size.} Since BRepCLIP uses cross-modal contrastive learning, larger batches enlarge the in-batch negative pool and improve alignment quality. As shown in Table~\ref{tab:ablation_batch_size}, increasing the batch size from 128 to 200 yields substantial gains of 172.7\% and 68.5\% in Top-1 and Top-20 accuracy, respectively, whereas further increasing it to 400 brings only marginal improvements of 0.23\% and 0.02\%. We therefore adopt a batch size of 200, which achieves near-identical performance to 400 while requiring roughly half the GPU memory, about $\sim$30 GB compared with $\sim$55 GB, consistent with findings in OpenShape~\cite{liu2023openshape}.

% \textbf{Batch size.} We further analyze the effect of contrastive batch size on BRepCLIP training. Since BRepCLIP is optimized with cross-modal contrastive learning, increasing the batch size enlarges the pool of in-batch negatives and can improve alignment quality. As shown in Table~\ref{tab:ablation_batch_size}, increasing the batch size from 128 to 200 yields a substantial improvement, boosting Top-1 retrieval by 172.7\% and Top-20 retrieval by 68.5\%. In contrast, increasing the batch size further from 200 to 400 brings only marginal gains of 0.23\% on Top-1 and 0.02\% on Top-20. We therefore adopt a batch size of 200 in the main experiments, as it achieves nearly the same retrieval performance as 400, with only a 0.23\% decrease in Top-1 and a 0.02\% decrease in Top-20, while requiring substantially less GPU memory. This observation is consistent with OpenShape~\cite{liu2023openshape}, which also finds that larger contrastive batches are beneficial, but that a medium batch size offers the best tradeoff between performance and training efficiency once returns begin to saturate.

\begin{wraptable}{r}{0.44\textwidth}
\vspace{-12pt}
\def\arraystretch{1.15}
\centering
\caption{Ablation of multimodal supervision on ABC retrieval.}
\vspace{4pt}
\label{tab:ablation_multimodal}
\resizebox{\linewidth}{!}{
\begin{tabular}{c c c|cccc}
\hline
\textbf{BRep} & \textbf{Image} & \textbf{MultiView}
& \textbf{Top-1} & \textbf{Top-5} & \textbf{Top-10} & \textbf{Top-20} \\
\hline
\textcolor{darkgreen}{\checkmark} & \textcolor{red}{\xmark} & \textcolor{red}{\xmark}         & 4.30 & 16.30 & 24.70 & 33.90 \\
\textcolor{darkgreen}{\checkmark} & \textcolor{darkgreen}{\checkmark} & \textcolor{red}{\xmark}  &
6.64 & 20.78 & 31.36 & 42.73 \\
\textcolor{darkgreen}{\checkmark} & \textcolor{darkgreen}{\checkmark} & \textcolor{darkgreen}{\checkmark} & \textbf{8.59} & \textbf{24.52} & \textbf{35.08} & \textbf{47.89} \\
\hline
\end{tabular}
}
\end{wraptable}

\noindent\textbf{Multimodal Supervision.} As shown in Table~\ref{tab:ablation_multimodal}, BRep-only training already provides a strong retrieval baseline. Adding single-view image supervision improves Top-1 and Top-20 by 54.4\% and 26.0\% respectively, confirming that visual supervision is complementary to native BRep geometry. Replacing single-view with multi-view supervision yields further gains of 29.4\% on Top-1 and 12.1\% on Top-20, indicating that richer visual coverage strengthens alignment between BRep structure and image-text semantics.

% \textbf{Multimodal supervision.} Table~\ref{tab:ablation_multimodal} shows that BRep-only training already yields a strong CAD-aware retrieval baseline. Adding single-view image supervision improves Top-1 retrieval by 54.4\% and Top-20 by 26.0\%, showing that visual supervision is complementary to native BRep geometry. Replacing single-view with multi-view image supervision brings a further 29.4\% gain in Top-1 and 12.1\% in Top-20, giving the best overall performance. This indicates that richer visual coverage improves the alignment between BRep structure and image-text semantics.
\section{Limitation}

BRepCLIP has two main limitations. First, faces and edges are tokenized at a fixed geometric resolution, which may be insufficient for complex CAD models with finer local detail or denser primitive counts, increasing memory and compute at scale. Second, semantic descriptors are limited to a fixed taxonomy of face and edge types, which does not cover the full diversity of primitives and topology encountered in real-world engineering data. Extending both the resolution and the semantic vocabulary are important directions for future work.

% BRepCLIP has two main limitations. First, it relies on fixed sampled point sets for faces and edges, with tokenizers trained at a fixed geometric resolution. Although sufficient for our current benchmarks, more complex CAD models with finer local detail or many more primitives may require denser sampling and longer sequences, increasing memory and compute cost. Second, the current semantic descriptors are limited to a fixed set of extracted face and edge types. While already useful for CAD-aware representation learning, they do not cover the full diversity of CAD primitives and topology encountered in real-world engineering data. Extending this taxonomy is an important direction for future work.

\section{Conclusion}

We presented BRepCLIP, the first multimodal contrastive pretraining framework built directly on BRep primitives for CAD understanding. By modeling faces and edges as distinct geometric entities, learning separate discrete token vocabularies for surface and curve geometry, and aligning the resulting BRep representation with text and image embeddings, BRepCLIP captures fine-grained CAD semantics that are typically lost in point-based representations. Across zero-shot text-to-CAD retrieval and zero-shot CAD classification, BRepCLIP consistently outperforms generic point-based encoders and strong multimodal baselines. We further showed that the learned embedding supports CAD-aware generation evaluation through BRepCLIP-Score, providing a more structure-sensitive alternative to image-based similarity metrics such as CLIP-Score. These results establish native BRep structure as a strong foundation for multimodal CAD representation learning, and open a new direction toward BRep-native foundation models for retrieval, evaluation, and broader engineering design workflows.

\section{Acknowledgements}
This work was co-funded by the European Union under Horizon Europe, grant number 101135724, project LUMINOUS. However, the views and opinions expressed are those of the author(s) only and do not necessarily reflect those of the European Union. Neither the European Union nor the granting authority can be held responsible. 

% \newpage
% \pagebreak
% \bibliographystyle{plainnat}
% \bibliography{references}

{
\bibliographystyle{plain} % or ieeenat_fullname
\bibliography{references}
}

% FOR SUPPLEMENTARY
\newpage

\appendix
\section*{Supplementary Material}

\section{Dataset Analysis}

In this section, we provide additional analysis of the datasets used for training and evaluation. Our training data is built from the high-quality ABC subset of CADCap-1M, from which we use 400K CAD models for training and 10K for validation. Retrieval is evaluated on a held-out ABC split with 91K samples, while zero-shot retrieval is evaluated on two unseen CAD datasets: Automate and CADParser. For zero-shot classification, we use FabWave. The original FabWave manifest contains 45 categories, but after filtering 43 broken or incomplete assets, the final benchmark contains 4,378 valid CAD models across 39 categories.

\subsection{Training Data Statistics}

\begin{figure}[ht]
    \centering
    \includegraphics[width=1\linewidth]{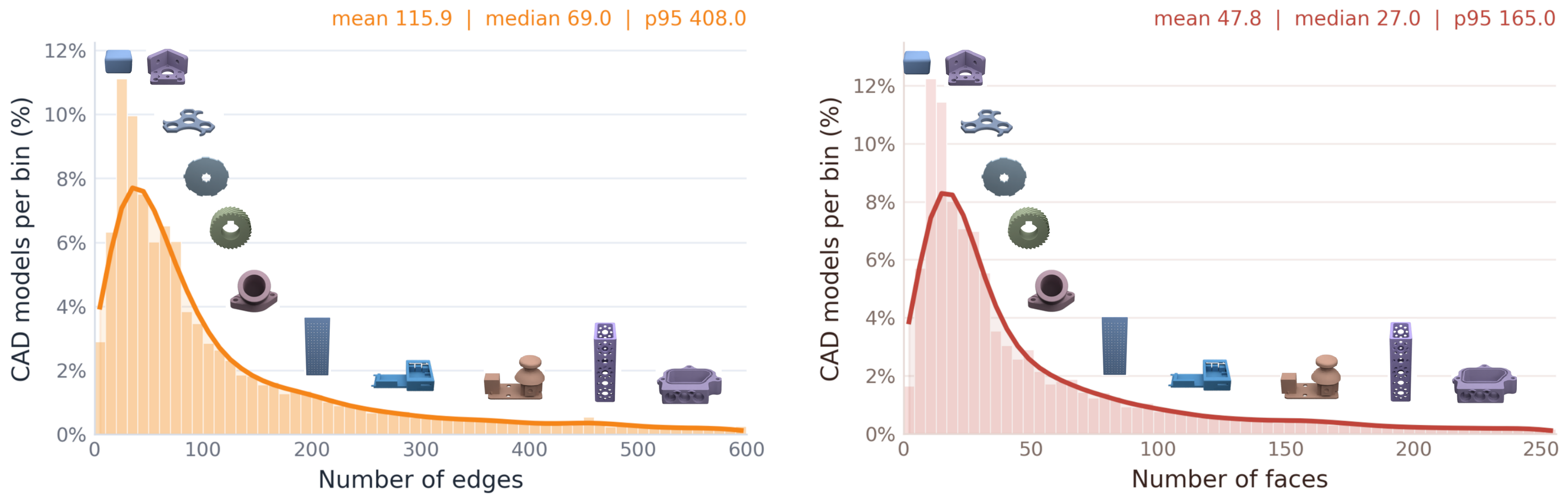}
    \caption{Distributions of the number of edges per CAD model (left), the number of faces per CAD model (middle), and the average number of edges per face (right) in the 400K ABC training set.}
    \label{fig:supp_dataset_edges_per_face}
\end{figure}

We first analyze the geometric complexity of the 400K ABC training set used for BRepCLIP pretraining. Figure~\ref{fig:supp_dataset_edges_per_face} summarizes three complementary statistics: the number of edges per CAD model, the number of faces per CAD model, and the average number of edges per face. All three provide a compact view of the structural diversity present in the training set.

The face and edge distributions are both strongly right-skewed, showing that most CAD models contain a relatively small to moderate number of primitives, while a smaller subset contains substantially more complex geometry. For faces, the mean number per CAD model is 47.8, the median is 27.0, and the 95th percentile is 165.0. This indicates that the training set contains many simple and medium-complexity mechanical parts, but also a substantial long tail of models with rich surface decomposition. For edges, the distribution is broader and heavier-tailed, with a mean of 115.9, a median of 69.0, and a 95th percentile of 408.0. This is expected, since edges capture local boundaries, transitions, and fine geometric details more densely than faces. The much heavier tail in the edge distribution confirms that many CAD models contain rich boundary structure, which motivates treating edges as first-class primitives rather than relying only on surface-level information.

To further characterize local BRep structure, Figure~\ref{fig:supp_dataset_edges_per_face} also plots the distribution of the average number of edges per face. This distribution is concentrated around 2.5, with a mean of 2.5, a median of 2.5, and a 95th percentile of 3.0. This indicates that most faces in the training set are bounded by a small number of edges, reflecting the predominance of regular engineering surfaces such as planar, cylindrical, and smoothly connected analytic patches. At the same time, the spread toward higher values suggests the presence of more irregular or highly segmented face boundaries in complex models.

Taken together, these statistics show that the ABC training split spans a broad range of CAD complexity, from simple low-face parts to highly structured objects with many faces and edges. This diversity is important for training BRepCLIP, since it exposes the model to both regular mechanical primitives and harder long-tail geometries.

\subsection{Evaluation Split Overview}

\begin{wrapfigure}{r}{0.58\textwidth}
\vspace{-10pt}
\centering
\includegraphics[width=\linewidth]{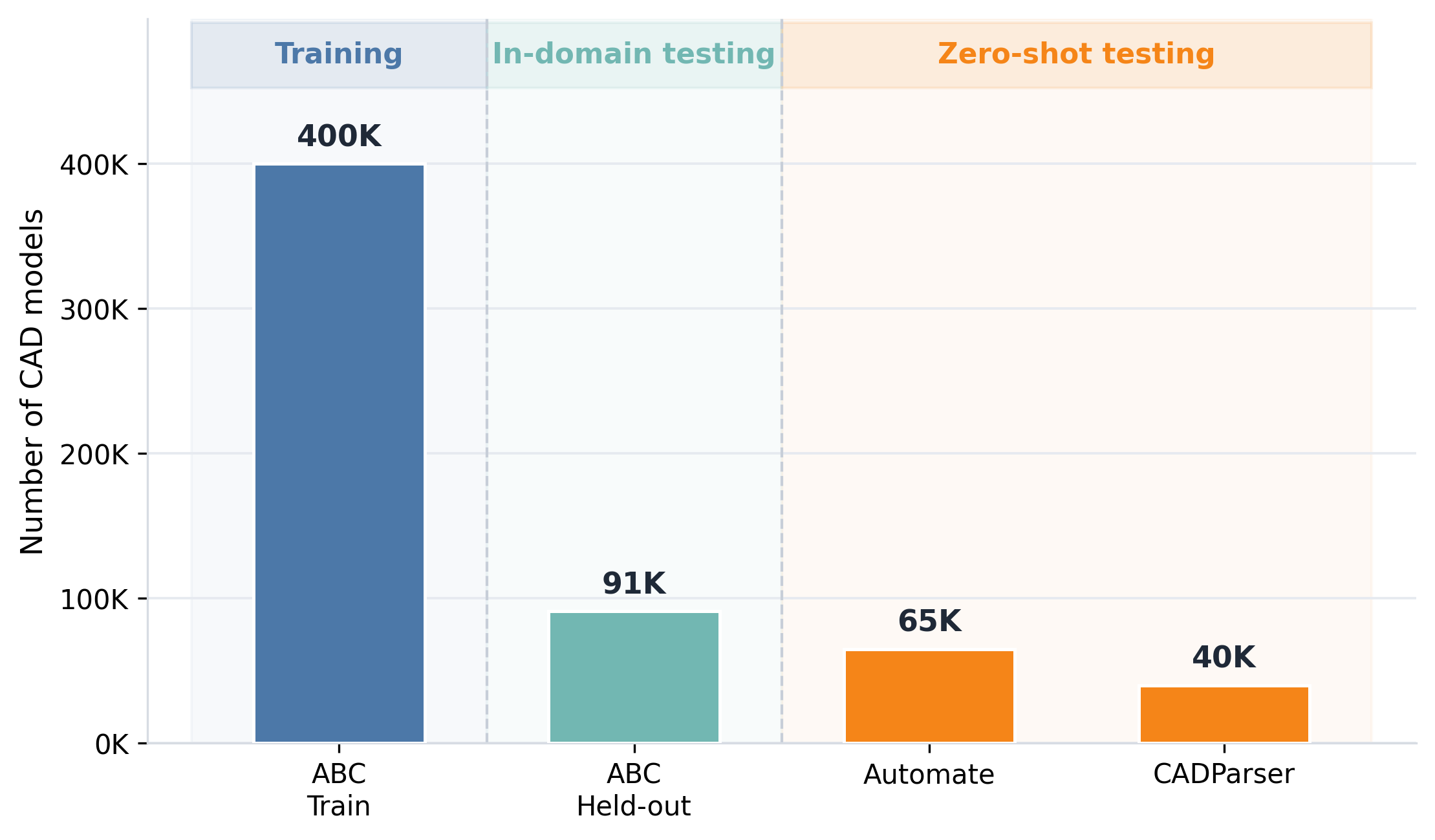}
\caption{Overview of training, in-domain retrieval, and zero-shot retrieval splits used in our experiments.}
\label{fig:supp_dataset_splits}
\vspace{-10pt}
\end{wrapfigure}
Figure~\ref{fig:supp_dataset_splits} summarizes the data usage across training and evaluation. The main training set consists of 400K ABC models. For in-domain retrieval testing, we use a held-out ABC split of 91K CAD models. For zero-shot retrieval transfer, we evaluate on two unseen datasets: Automate with 65K models and CADParser with 40K models. This setup clearly separates in-domain retrieval from zero-shot transfer evaluation, allowing us to test both memorization-free retrieval within the same CAD source and generalization to different CAD repositories.

For zero-shot classification, we use FabWave after filtering invalid assets. The final benchmark contains 4,378 valid samples across 39 categories and is never used during training. This makes FabWave a strict zero-shot transfer benchmark for category-level CAD recognition.

\subsection{Primitive Type Statistics}

We further analyze the distribution of BRep primitive types in the 400K ABC training set. Our extraction pipeline assigns semantic labels to both faces and edges. For faces, we extract \textit{Plane}, \textit{Cylinder}, \textit{Cone}, \textit{Sphere}, \textit{Torus}, and \textit{Rational NURBS}. For edges, we extract \textit{Line}, \textit{Circle}, \textit{Ellipse}, \textit{Non-rational B-spline}, and \textit{Rational B-spline}. We also extract edge relation attributes, including \textit{Convex}, \textit{Concave}, \textit{Smooth}, and \textit{Closed}.

\begin{figure}[ht]
    \centering
    \includegraphics[width=1\linewidth]{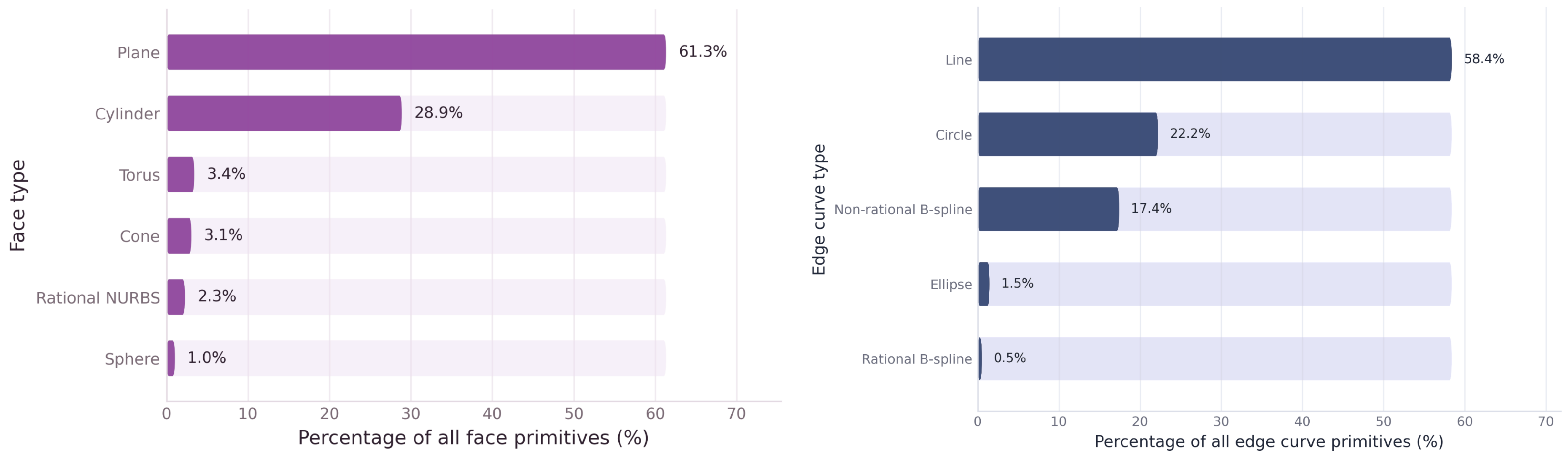}
    \caption{Distribution of face primitive types (left) and edge curve types (right) in the 400K ABC training set.}
    \label{fig:supp_primitive_types}
\end{figure}

\begin{wrapfigure}{r}{0.52\textwidth}
\vspace{-8pt}
    \centering
    \includegraphics[width=\linewidth]{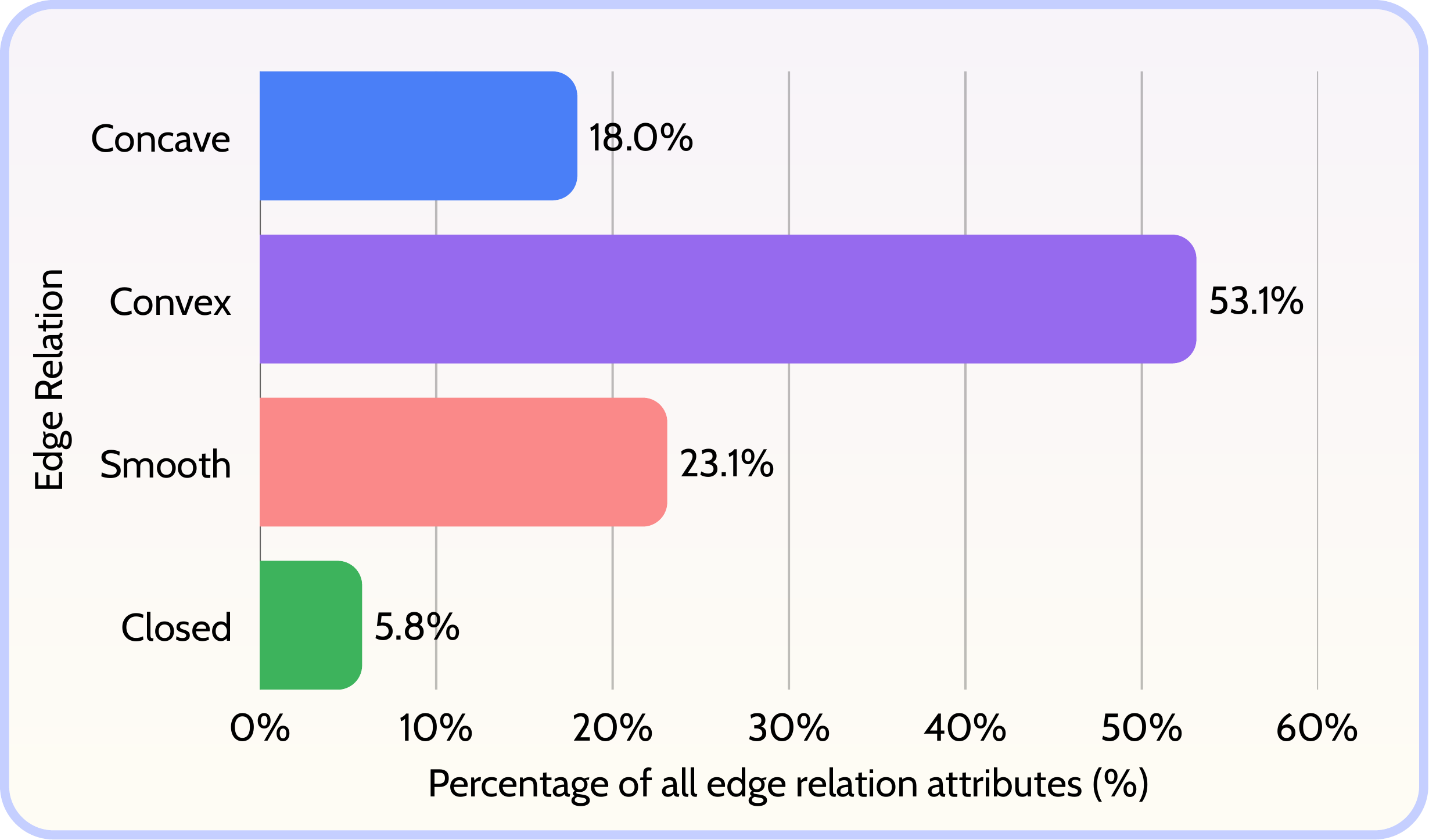}
    \caption{Distribution of edge relation attributes in the 400K ABC training set.}
    \label{fig:supp_edge_relations}
\vspace{-8pt}
\end{wrapfigure}
Figure~\ref{fig:supp_primitive_types} shows that the dataset is dominated by analytic CAD geometry. For faces, planes account for 61.3\% of all face primitives, followed by cylinders at 28.9\%. Torus, cone, rational NURBS, and sphere faces are much less frequent. For edges, lines are the most common primitive at 58.4\%, followed by circles at 22.2\% and non-rational B-splines at 17.4\%, while ellipses and rational B-splines are relatively rare. Overall, this confirms that most CAD models in ABC are composed of planar and cylindrical surfaces bounded by straight and circular edges, with a smaller long tail of more complex free-form geometry.

Figure~\ref{fig:supp_edge_relations} shows the distribution of edge relation attributes. Convex edges are the most common at 53.1\%, followed by smooth edges at 23.1\% and concave edges at 18.0\%. Closed edges account for the remaining 5.8\%. This indicates that most CAD parts are dominated by regular outward boundaries and smooth transitions, while concave and closed structures occur less frequently but remain important for engineering geometry.

Taken together, these primitive statistics support our use of primitive-aware tokenization and semantic descriptors, since face and edge types provide useful structural cues beyond raw point samples alone.

\begin{figure}[t]
    \centering
    \includegraphics[width=1\linewidth]{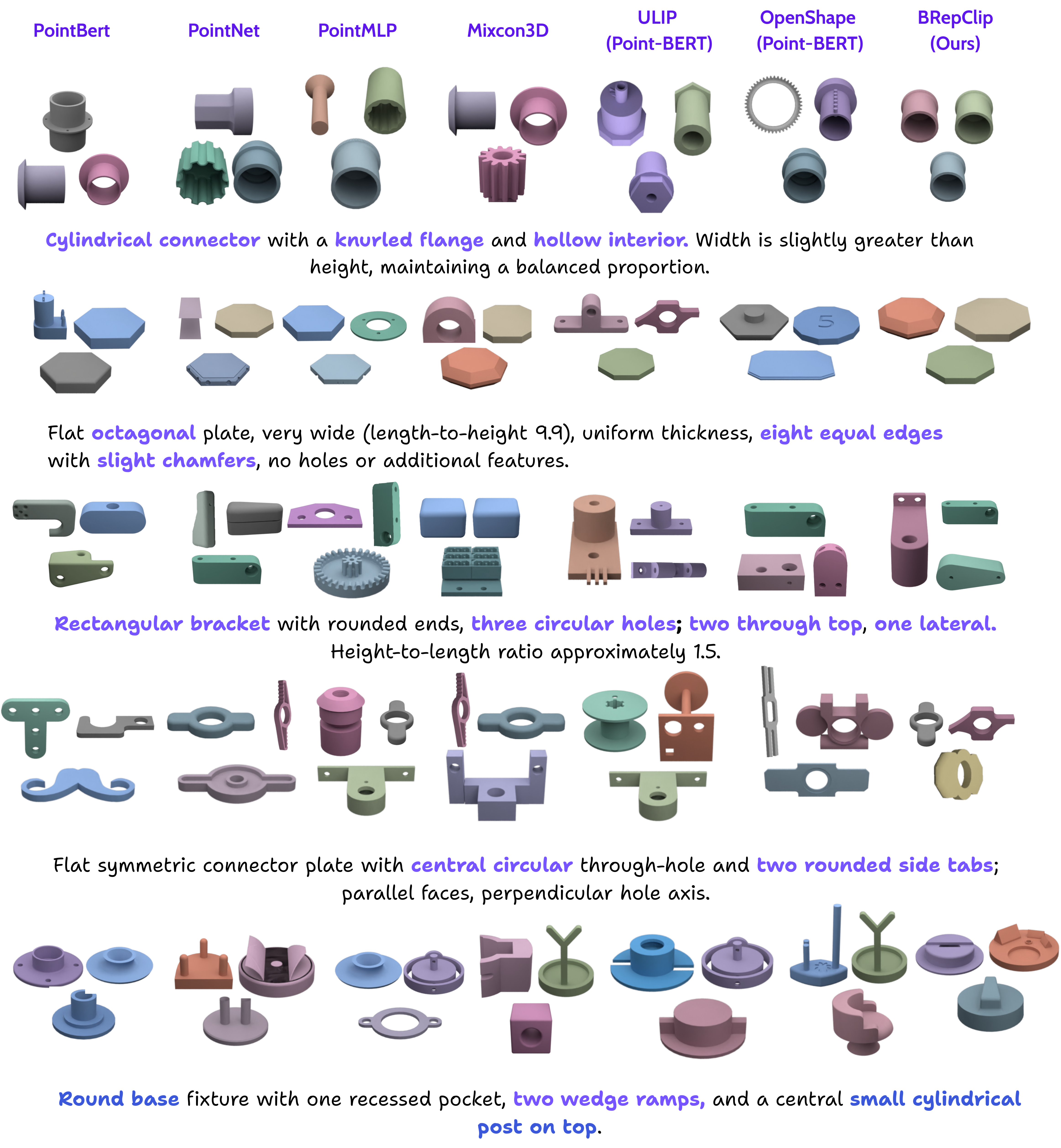}
    \caption{Additional qualitative text-to-CAD retrieval results. Given a text query, BRepCLIP retrieves CAD models that better preserve fine-grained geometric details such as hole layout, boundary structure, and surface composition than point-based baselines.}
    \label{fig:supp-qualitative}
\end{figure}

\begin{figure}[t]
    \centering
    \includegraphics[width=1\linewidth]{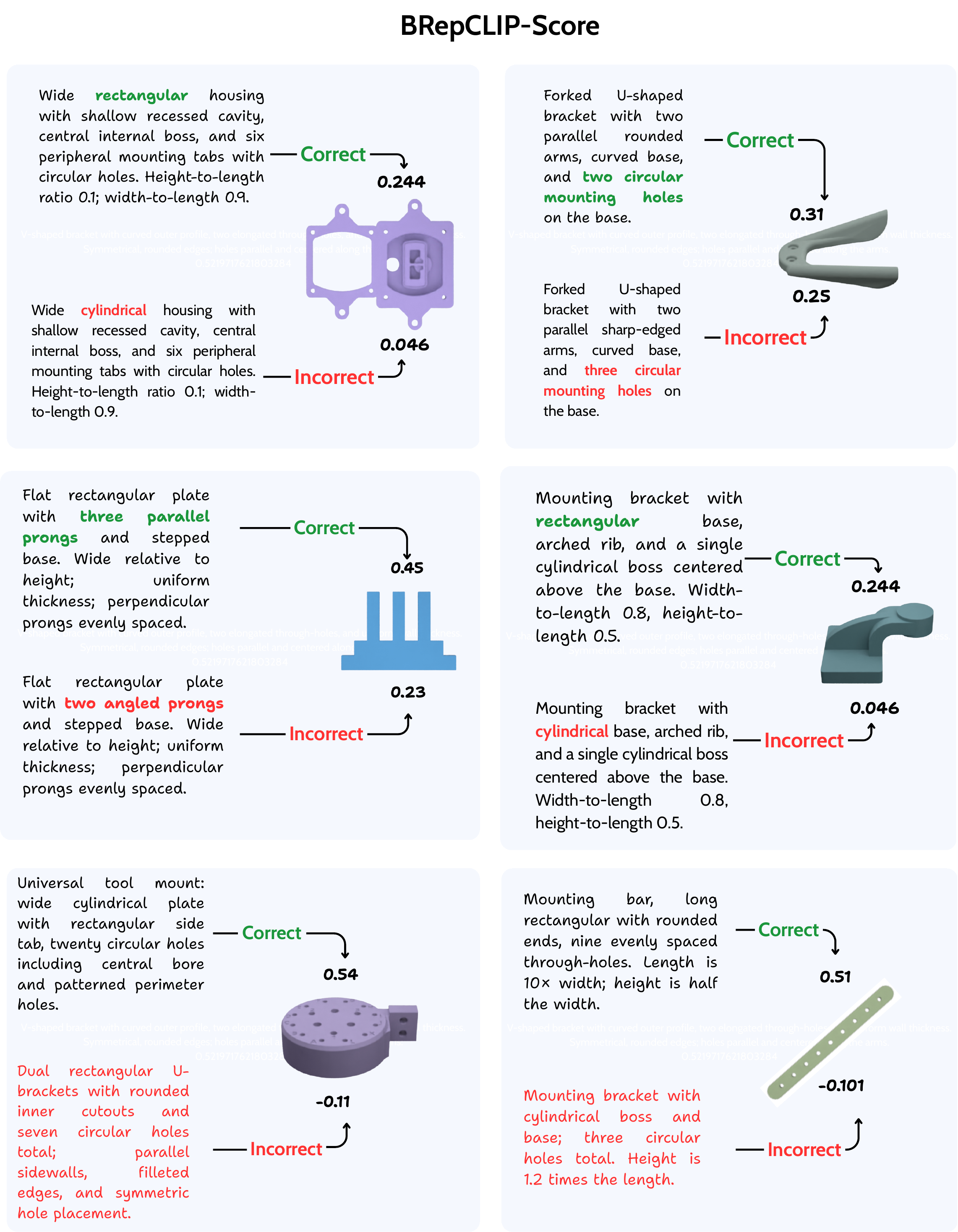}
    \caption{Additional qualitative results for BRepCLIP-Score. Higher scores are assigned to CAD models that better match the input text in geometry and structure, while semantically inconsistent generations receive lower scores.}
    \label{fig:supp-brepclip}
\end{figure}

\begin{figure}[t]
    \centering    \includegraphics[width=1\linewidth]{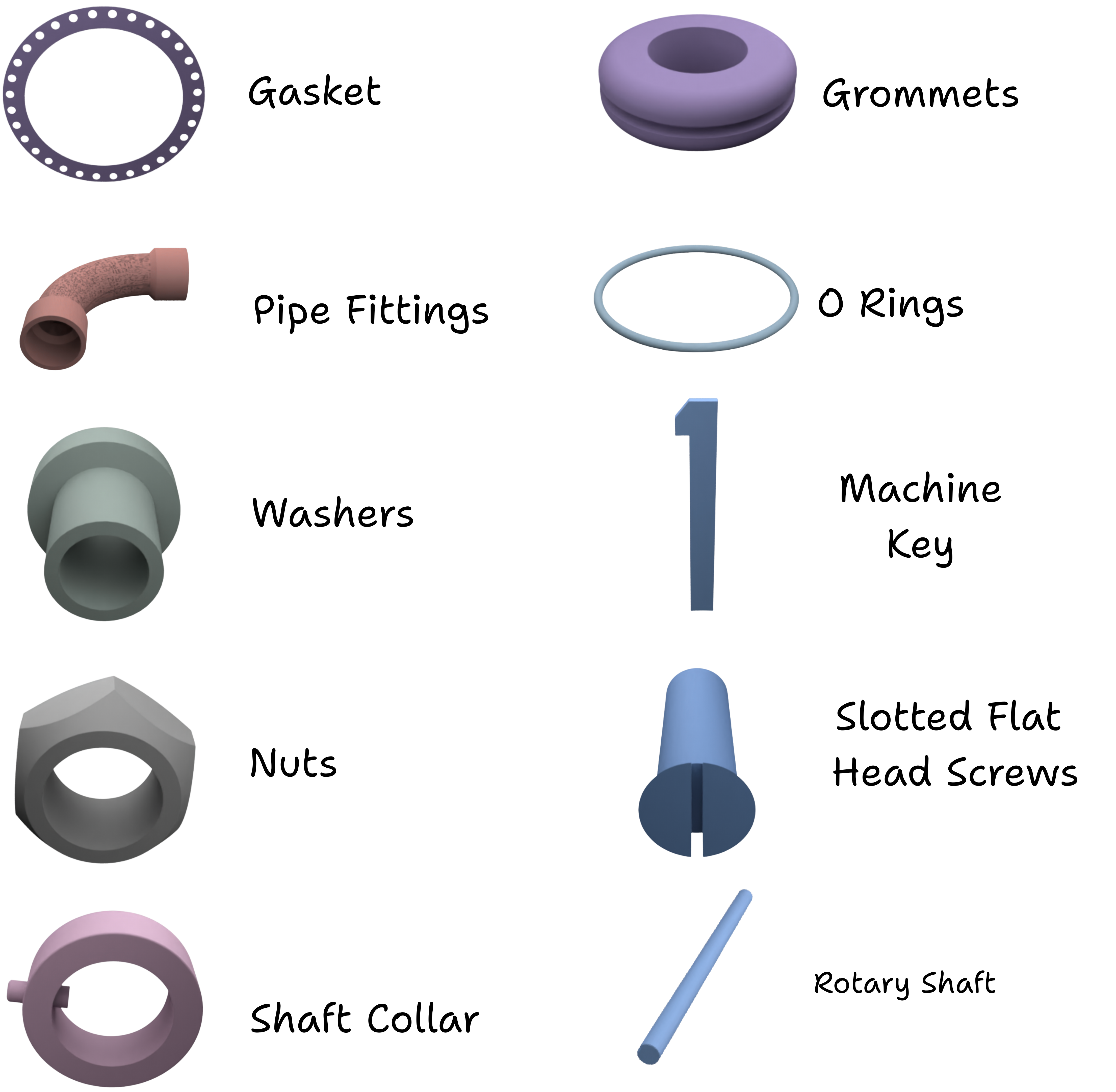}
    \caption{Additional qualitative results for zero-shot classification on FabWave. BRepCLIP produces more semantically accurate class predictions for engineering CAD models than point-based and multimodal baselines.}
    \label{fig:supp-zero-shot}
\end{figure}

\textbf{More Qualitative results}. In Figure~\ref{fig:supp-qualitative},~\ref{fig:supp-brepclip},~\ref{fig:supp-zero-shot} we provided more qualitative samples on retrieval task, BRepCLIP-Score and zero-shot classification.

% \subsection{More results on Zero-Shot Classification}

% \textbf{More results on BRepCLIP Score} Figure~\ref{fig:supp-brepclip} shows more qualitative results on BRepCLIP-Score.

% \clearpage
% \newpage

% \input{checklist}
\end{document}